\documentclass[conference]{IEEEtran}

\usepackage{comment}
\usepackage{textgreek}
\usepackage{float}
\usepackage{url}
\usepackage{booktabs}
\usepackage{siunitx}
\usepackage{graphicx}
\usepackage{array}
\usepackage{makecell}
\usepackage{pdflscape}
\usepackage{amsmath,amssymb,amsfonts}
\usepackage{algorithmic}
\usepackage{textcomp}
\usepackage{xcolor}
\usepackage[utf8]{inputenc}
\usepackage{graphicx}
\usepackage{amsmath}
\usepackage{longtable}
\usepackage{tabularx}
\usepackage{rotating}
\usepackage{array}
\renewcommand{\arraystretch}{1.5}
\usepackage{threeparttable}
\usepackage{stfloats}
\usepackage{wrapfig}
\usepackage{hyperref}
\usepackage{caption}

\graphicspath{{figures/}{pictures/}{images/}{./}} 

\usepackage{times}                     

\usepackage{booktabs}                  
\usepackage{lipsum}                    
\usepackage{mwe}                       
    \renewcommand{\arraystretch}{1.5} 

\ifCLASSINFOpdf

\else

\fi

\begin{document}
\pagestyle{plain}

\title{Evaluating Magic Leap 2 Tool Tracking for AR Sensor Guidance in Industrial Inspections}

\author{\IEEEauthorblockN{Christian Masuhr, Julian Koch and Thorsten Schüppstuhl}
\IEEEauthorblockA{Institute of Aircraft Production Technology\\
Hamburg University of Technology (TUHH)\\
Hamburg, Germany\\
Email: \{christian.masuhr, julian.koch, schueppstuhl\}@tuhh.de}}

\maketitle

\begin{figure*}[!t]
    \centering
    \includegraphics[width=1\textwidth]{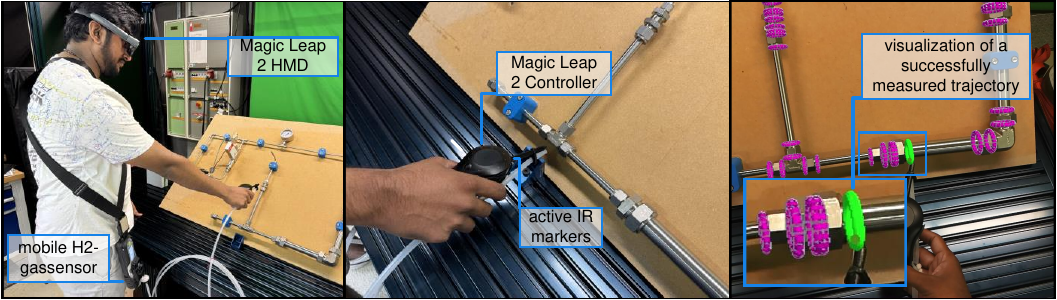}
    \caption{Overview of AR assisted leakage inspection.}
    \label{fig:leakage_inspection}
\end{figure*}

\begin{abstract}
Rigorous evaluation of commercial Augmented Reality (AR) hardware is crucial, yet public benchmarks for tool tracking on modern Head-Mounted Displays (HMDs) are limited. This paper addresses this gap by systematically assessing the Magic Leap 2 (ML2) controller’s tracking performance. Using a robotic arm for repeatable motion (EN ISO 9283) and an optical tracking system as ground truth, our protocol evaluates static and dynamic performance under various conditions, including realistic paths from a hydrogen leak inspection use case. The results provide a quantitative baseline of the ML2 controller's accuracy and repeatability and present a robust, transferable evaluation methodology. The findings provide a basis to assess the controller’s suitability for the inspection use case and similar industrial sensor-based AR guidance tasks.
\end{abstract}

\IEEEpeerreviewmaketitle

\section{Introduction}

Many safety-critical industrial tasks, like non-destructive testing (NDT), still rely on manual sensor inspections prone to human error, which compromises data quality and safety \cite{Bertovic2015, Torres2021, Koch2024}. Augmented Reality (AR) can enhance these procedures by superimposing digital guidance---such as optimal trajectories or sensor data---directly onto the operator's view, improving accuracy and efficiency \cite{Voinea2023, Marino2021, Frandsen2022, WANG2016406}. This approach is particularly promising for tasks involving handheld sensors, where adherence to guidance protocols is critical for data integrity \cite{polvi2018, Koch2023}. This work therefore focuses on AR-assisted monitoring of manual sensor trajectories.

A highly relevant use case is the manual leak inspection on hydrogen electrolyzers (see Figure~\ref{fig:leakage_inspection}). As a prominent but volatile energy carrier, hydrogen handling is safety-critical and requires guiding a sensor along predefined paths with precise control over speed and distance \cite{Akpasi2025, Masuhr2024}. Current ad-hoc inspection practices are complicated by challenges like invisibility of gas and sensor delays. These demanding conditions establish hydrogen leak inspection as an ideal testbed for evaluating AR sensor guidance \cite{TCHOUVELEV202112439}.

The effectiveness of AR guidance hinges on stable object registration and precise tool tracking \cite{Azuma1997}. While specialized systems for AR tool tracking can achieve high accuracy, their integration overhead often limits practical deployment. Commercial off-the-shelf systems like the Magic Leap 2 (ML2) feature a distinct controller with its own independent simultaneous localization and mapping (SLAM) system. These offer a compelling alternative but lack public performance benchmarks. This creates a critical knowledge gap for practitioners assessing the trade-off between accuracy and effort. 

This study is motivated by previous work on the HoloLens 2, which revealed limitations in existing systems \cite{MasuhrUsability}, and by preliminary tests showing promising static accuracy of 1.39 mm for the ML2 controller. A review of current literature highlights two major gaps. First, there is no quantitative benchmarking of the ML2's performance under challenging industrial conditions, such as limited accessibility or reflective surfaces. Second, many published tracking studies lack methodological rigor, often relying on non-repeatable manual movements and insufficient documentation for reproducibility or comparison.

This paper addresses these gaps by presenting a systematic evaluation of the static and dynamic pose tracking accuracy of the ML2 controller. Using a robot-driven approach that combines standardized procedures with application-specific trajectories, it provides the first comprehensive benchmark of the controller’s performance for safety-critical industrial tasks, with an optical tracking system (OptiTrack) serving as ground truth.

While this work is validated within the specific use case of hydrogen leak inspection, the underlying methods and findings are transferable to other industrial applications that require precise, AR-assisted manual sensor guidance. 

The remainder of this paper is structured as follows: Section 2 reviews related work, Section 3 details the methodology, Section 4 presents the results, Section 5 discusses their implications, and Section 6 provides a conclusion and outlook.

\section{STATE OF THE ART}
This section reviews existing technologies for AR tool tracking and critically examines the methodologies used in prior work to evaluate their accuracy, providing the context for this paper's contribution.

\subsection{Tool Tracking Technologies for AR in Industrial Settings}
Effective industrial AR hinges on accurate tool tracking. A primary distinction exists between outside-in systems (e.g. OptiTrack and Vicon), which use fixed external sensors to offer high precision but limited mobility. This makes them less suitable for inspections in large workspaces \cite{Nancel2021HL2ViveComp}. In contrast, inside-out systems, integrated into devices like the ML2, use onboard sensors (cameras, IMUs) and SLAM or visual-inertial odometry (VIO) algorithms to determine their own pose, offering the crucial mobility for industrial applications \cite{Voinea2023}\cite{Campos2021ORBSLAM3}.

To achieve tool tracking itself, approaches include marker-based tracking with physical references \cite{MartinGomez2024STTAR, Brand2020} and model-based tracking. The approach investigated in this work leverages dedicated handheld controllers with their own integrated tracking systems. These often provide higher reliability than general-purpose hand tracking \cite{Soares2021HL2HandTracking} and, as exemplified by the ML2 controller, offer a commercially available, inside-out solution with minimal setup effort.

\subsection{Accuracy Analysis of Tool Tracking Systems}
The accuracy of AR tracking systems varies widely, necessitating device- and application-specific evaluation. While medical applications have demonstrated sub-millimeter accuracies for the HoloLens 2 using custom marker setups \cite{MartinGomez2024STTAR, Gsaxner2021, Mai2022}, these results are not directly transferable to industrial scenarios with challenging surfaces.

Despite these data points, a significant research gap exists for the markerless, inside-out controller tracking of enterprise-grade devices. While the ML2's general head-tracking accuracy has been validated \cite{GaitBalance2024, VanDoorn2025}, a systematic, quantitative evaluation of its controller performance—especially under the unique challenges of industrial tasks like hydrogen leak inspection with reflective components—is missing from the literature.

The methodology for rigorous analyses has also evolved. Many studies have been limited by reliance on non-repeatable manual movements and a focus on application-specific trajectories \cite{Boal2023, Costa2024, Hu2022}, a known limitation for comparability \cite{Gsaxner2021}. Consequently, recent state-of-the-art investigations increasingly employ robotic arms to create repeatable and standardized test conditions \cite{MartinGomez2024STTAR, Passos2020, Sitole2020}.

Lacking a single, universally accepted evaluation standard \cite{Luckett2019}, state-of-the-art assessments must create a comprehensive evaluation by adapting procedures from multiple domains \cite{Son2022, Marino2022Benchmarking}. Frameworks are often adapted from industrial robotics (e.g., EN ISO 9283 \cite{en_iso_9283}) and supplemented with metrics from SLAM research (e.g., ATE \cite{Sturm2012}). Key stability indicators like jitter, drift, and latency are also essential for usability \cite{Xu2025}, and a high-precision ground truth system remains a fundamental prerequisite \cite{Brand2020, Sa2022}.

\section{METHODOLOGIES}
This section details the requirements derived from the use case, the experimental setup, and the methodologies employed for evaluating the tracking performance of the ML2 controller.

\subsection{Use-Case of an AR-Assisted Leakage Inspection}
\begin{figure*}[!t]
    \centering
    \includegraphics[width=1\linewidth]{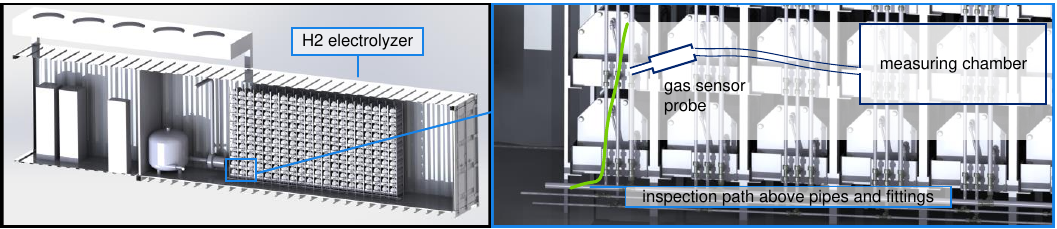}
    \caption{Electrolyzer leakage inspection.}
    \label{fig:electrolyzer}
\end{figure*}

Accurate leak localization requires high-fidelity guidance of a gas sniffer sensor, maintaining a consistent trajectory, velocity, and proximity while navigating complex electrolyzer geometries \cite{Masuhr2024}. Environmental challenges like reflective, texture-poor metallic surfaces can impair optical sensing (see Figure \ref{fig:electrolyzer}). Given that H$_{2}$ components are often designed in 1/4-inch dimensions, stringent spatial accuracy is required for reliable gas detection. Studies confirm that high translational accuracy is paramount for detecting small leaks, while rotational accuracy, though less stringent, remains significant. Based on preliminary tests, we establish a primary end-to-end requirement of \textless 5 mm and \textless 10$^{\circ}$ deviation, a metric that encompasses both tracking precision and object registration.

The operational environment imposes further demands. The dense layout of electrolyzer systems leads to frequent line-of-sight occlusions, necessitating a tracking system that is robust to interruptions and capable of rapid re-acquisition. Furthermore, temporal stability, which is characterized by minimal jitter and low long-term drift, is critical for smooth and trustworthy user guidance. To assist the operator, the AR application visualizes target trajectories (e.g., around fittings, see Figure~\ref{fig:leakage_inspection}) and can apply data processing to compensate for the sensor's inherent signal delay.

\subsection{Rationale for Technology Selection}
\label{sec:RationaleTechSelection}

The ML2 was selected as the core platform due to several key advantages. Its controller, which runs its own independent SLAM system utilizing an IMU and cameras, provides real-time tool tracking via a pose stream and active IR markers, which, unlike passive markers, function without an external light source. Furthermore, an external compute unit (Power Puck) ensures the headset remains lightweight and ergonomic for extended use. These features, combined with its processing power and developer community support, make it a suitable platform for this work.

\begin{figure}
    \centering
    \includegraphics[width=1\linewidth]{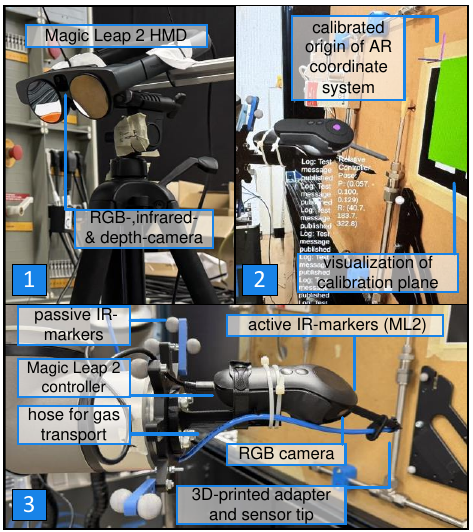}
    \caption{ML2 with tripod mount (1); ML2 controller attached to UR10e robot (2); Camera capture of the running data capture application (3).}
    \label{fig:Closeup-Magic Leap 2}
\end{figure}

\subsection{Experimental Methodology}
\label{Sec:Experimental Methodology}
This study quantified the ML2 controller's static and dynamic tracking accuracy and repeatability using a high-precision hardware setup to establish a performance baseline for industrial AR guidance.

\subsubsection{Hardware and Experimental Setup}
\begin{figure}
    \centering
    \includegraphics[width=1\linewidth]{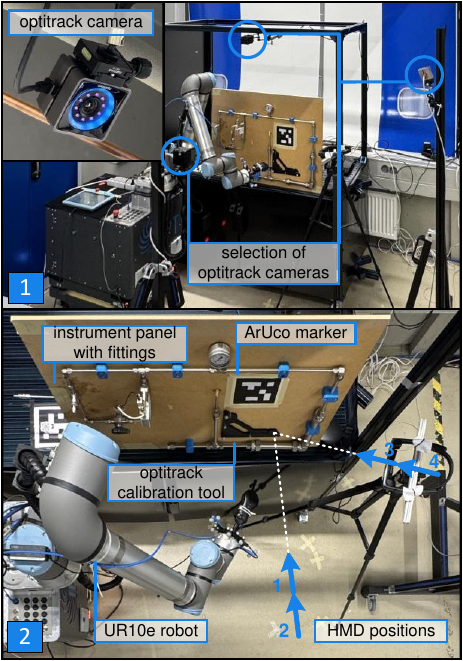}
    \caption{Part of the camera configuration of the used Optitrack system (1); Hardware setup for the ML2 tool tracking accuracy analysis on an instrument panel (2).}
    \label{fig:setup_study_1}
\end{figure}

The experimental setup, shown in Figure~\ref{fig:setup_study_1}, was designed to simulate a realistic industrial inspection environment, with  stainless steel pipes, various flanges, and fittings. To ensure precise and repeatable movements, a Universal Robots UR10e robotic arm (manufacturer-reported repeatability: $\pm0.05$\,mm) was used to manipulate the ML2 controller. The controller was attached to the robot's end-effector via a custom 3D-printed fixture, which incorporated nine asymmetrically placed infrared (IR) markers  (see Figure~\ref{fig:Closeup-Magic Leap 2}). To integrate the gas sensor, a second custom 3D-printed attachment was designed to mount the sensor probe onto the controller, ensuring clearance for its tracking sensors (see Figure~\ref{fig:Closeup-Magic Leap 2}~(2)). This setup allows the controller’s real-time pose to approximate the sensor tip’s pose after a one-time calibration. An OptiTrack motion capture system with six Prime X13 cameras, providing a reported accuracy of $\pm0.2$\,mm (see Figure \ref{fig:setup_study_1} (1)). The Unity-based AR application used a Node-RED backend with a local MQTT server to control experiment logic and log data to .csv files.

\subsubsection{Coordinate System Alignment and Data Capture}
A crucial prerequisite for accurate measurement is the precise alignment of all coordinate systems. The OptiTrack system was calibrated once using a standard optitrack calibration tool to establish a stable ground truth coordinate system within the workspace (see Figure \ref{fig:setup_study_1} (2)). To align the AR system's world space with this ground truth, an ArUco marker was placed on the instrument panel \cite{GarridoJurado2014}. As the headset's physical position was changed for the different test series (P1-P4), the registration of the AR coordinate system was re-initialized before each series by viewing the ArUco marker and set a fixed origin point. The application's user interface, depicted in Figure \ref{fig:Closeup-Magic Leap 2} (3), allowed for real-time monitoring and control of the experimental procedure. To account for the AR system's variable frame rate output, data from the ML2 and OptiTrack systems were captured with high-resolution timestamps and aligned in post-processing based on the nearest timestamp.

This re-calibration before each measurement run is analogous to real-world application workflows and serves a critical methodological purpose: it ensures that any potential long-term drift of the AR system's world anchor does not accumulate across different test conditions, thus isolating the analysis to the controller's tracking performance relative to its just-established origin. While this introduces a new calibration error for each run, it allows for an independent analysis of the system's precision (scatter) and its systematic offset (mean deviation), which are not directly correlated.

\subsubsection{Accuracy and Repeatability Tests:}
\begin{figure} [H]
    \centering
    \includegraphics[width=1\linewidth]{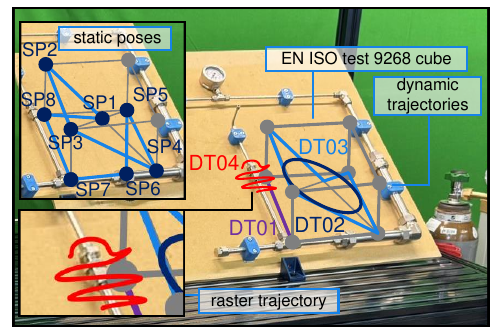}
    \caption{Poses and trajectories for experiment.}
    \label{fig:experiment_poses_trajecotries_accuracy}
\end{figure}

In the absence of a dedicated standardized test method, we developed a custom approach based on industrial robotics standards (EN ISO 9283) and guidelines for assessing localization and tracking systems (ISO/IEC 18305) \cite{ISO18305}. A structured experimental protocol was designed to evaluate the spatial tracking accuracy of the ML2 controller. The focus was on capturing both standard performance metrics for baseline benchmarking and application-specific indicators relevant to practical robustness and reliability (see Table \ref{tab:comprehensive_test_protocol_final}). These include trajectories specific to leakage inspection, as well as temporal tracking stability (jitter, drift) and resistance to occlusion \cite{Xu2025}. The protocol included static and dynamic test sequences to thoroughly assess tracking performance (see Figure~\ref{fig:experiment_poses_trajecotries_accuracy}). Test positions and trajectories were selected to align with both standard requirements and realistic interactions between the AR headset and the tracked tool during inspection tasks inside an H\textsubscript{2} electrolyzer. This combination strengthens the methodological design by ensuring that the evaluation is not only quantifiable but also representative of real-world constraints and operator-tool interactions. 

Three primary parameters were identified for investigation: sensor movement speed, and the position and orientation of the head-mounted display (HMD). To assess how variations in these parameters affect tracking accuracy, the entire test protocol was repeated for four distinct HMD positions and two movement speeds.

The HMD positions, constrained by the robot's working volume, simulated realistic user-tool configurations and included two frontal views at approx. 500 mm (F1) and 600 mm (F2), and two at a 45° lateral offset at approx. 400 mm (S3) and 500 mm (S4) (see Figure~\ref{fig:setup_study_1}~(2)). Dynamic trajectories were executed at two speeds: a practical baseline of 5 mm/s for detecting low leakage rates, and a more demanding 50 mm/s to investigate the system's performance limits, justified by the capabilities of modern sensors \cite{Inficion.RoboticSniffing}. Static tests were primarily conducted with an approach speed of 50 mm/s, while two positions were also tested at 25 mm/s to assess speed influence. Data was recorded at 10 Hz for the lower speed, and at 50 Hz (OptiTrack) and approx. 30 Hz (ML2) for the higher speed.

A total of eight static poses (SP01–SP08) were defined and executed, each repeated 30 times, to enable a statistically robust analysis of tracking repeatability and deviation. At each static pose, the robot held its position for 2.5 seconds before a measurement was recorded to reduce the effects of the robot’s residual motion. Four dynamic trajectories (DT01–DT04), including linear, circular, toroidal, and raster scan paths, were performed in 10 repetitions each, following the specifications of EN ISO 9283.

A scaled-down ISO test cube, in accordance with EN ISO 9283 but reduced to an edge length of 200 mm, was used to fit within the field of view of the tracking sensors (see Figure \ref{fig:experiment_poses_trajecotries_accuracy}). On this cube, the static pose targets were positioned on an inclined plane to simulate typical access angles. In addition, two standard trajectories for evaluating path accuracy (DT02 circle and DT03 square) were defined on this inclined surface. Another trajectory, executed directly above a pipe, was designed to reflect realistic operational conditions for leakage inspection (DT01 line). Finally, two distinct task-oriented motion paths were included: a toroidal path around a pipe fitting (DT04-1) and a raster scan to reflect a realistic inspection workflow (DT04-2). For practical efficiency, the parallel lines composing the raster scan were combined and executed as a single, continuous robot program.

\begin{table}[!h] 
\scriptsize
\centering
\caption{Experimental protocol overview.}
\label{tab:comprehensive_test_protocol_final}

\begin{tabularx}{\columnwidth}{|l|p{1.5cm}|X|c|}
\hline
\textbf{ID} & \textbf{Category} & \textbf{Description} & \textbf{n} \\
\hline \hline
\multicolumn{4}{|l|}{\textit{\textbf{Part 1: Static Poses (Accuracy \& Repeatability)}}} \\
\hline
SP01 & ISO 9283 & Center of ISO Test Cube & 30 \\
\hline
SP02 & ISO 9283 & Diagonal Point 1 of ISO Test Cube & 30 \\
\hline
SP03 & ISO 9283 & Diagonal Point 2 of ISO Test Cube & 30 \\
\hline
SP04 & ISO 9283 & Diagonal Point 3 of ISO Test Cube & 30 \\
\hline
SP05 & ISO 9283 & Diagonal Point 4 of ISO Test Cube & 30 \\
\hline
SP06 & App-Specific & Pose near Pipe Connector A & 30 \\
\hline
SP07 & App-Specific & Pose near Pipe Connector B & 30 \\
\hline
SP08 & App-Specific & Scan above Pipe C & 30 \\
\hline \hline
\multicolumn{4}{|l|}{\textit{\textbf{Part 2: Dynamic Trajectories (Path Accuracy)}}} \\
\hline
DT01 & ISO 9283 & Linear path (L=500\,mm), parallel to central panel region. & 10 \\
\hline
DT02 & ISO 9283 & Circular path (R=200\,mm), parallel to central panel. & 10 \\
\hline
DT03 & ISO 9283 & Square path (S=300\,mm) in inclined plane intersecting panel features. & 10 \\
\hline
DT04-1 & App-Specific & Torus path around Pipe D: R=100\,mm, r=30\,mm. & 10 \\
\hline
DT04-2 & App-Specific & Raster scan path over 10×10\,cm area above pipe with 5 parallel lines. & 10 \\
\hline \hline
\multicolumn{4}{|l|}{\textit{\textbf{Part 3: System-Level Tests (Robustness \& Reliability)}}} \\
\hline
RT03 & Occlusion & To assess recovery from signal loss, the controller at a static pose (e.g., SP06) was occluded by hand for 5\,s (first partially, then fully). & 10 \\
\hline
SYT01 & Reliability & To quantify usability, initialization time was measured from system start until the first stable pose was reported by the \,\,controller. & 10 \\
\hline
SYT02 & Reliability & To quantify usability, re-acquisition time was measured after moving the controller out of the tracking volume and back in until a stable track was re-established. & 10 \\
\hline
ST01 & Stability & To quantify long-term drift, the controller was held at a static pose (e.g., SP01) for 30–60 min while its position and orientation were continuously recorded. & 1 \\
\hline
\end{tabularx}
\end{table}

\subsection{Metrics and Data Evaluation}
\label{Metrics and Data Evaluation}

\begin{table} [t]
    \scriptsize
    \centering
    \caption{Standard performance metrics used to assess system accuracy, repeatability, robustness, and stability.}
    \label{tab:evaluation_metrics}
    \begin{tabularx}{\columnwidth}{|p{0.5cm}|p{1.7cm}|X|>{\centering\arraybackslash}p{0.9cm}|}
        \hline
        \textbf{ID} & \textbf{Category} & \textbf{Description} & \textbf{Unit} \\
        \hline \hline
        \multicolumn{4}{|l|}{\textit{\textbf{Accuracy \& Repeatability}}} \\
        \hline
        AP & Position \,Accuracy & Mean Euclidean distance between reported position and \,\,OptiTrack ground truth. & mm (1σ) \\
        \hline
        AO & Orientation Accuracy & Mean angular deviation from the reference orientation. & deg (1σ) \\
        \hline
        RP & Position \,Repeatability & Standard deviation of repeated measurements at a static pose. & mm (1σ) \\
        \hline
        STB & Jitter / Pose Stability & Short-term fluctuation of position and orientation while stationary (RMS) & mm / deg \\
        \hline
        TP & Trajectory Accuracy & Mean deviation from reference path over entire dynamic \,\,trajectory. & mm (1σ) \\
        \hline
        TDR & Trajectory Drift & Accumulated deviation from the expected trajectory over time, indicating drift during motion. & mm/s, mm/min \\
        \hline \hline
        \multicolumn{4}{|l|}{\textit{\textbf{Robustness}}} \\
        \hline
        OSR & Occlusion Success Rate & Percentage of successful tracking re-acquisitions after partial/full occlusion. & \% \\
        \hline
        ROT & Re-acquisition Time \,(Occlusion) & Time required to re-establish tracking after temporary \,\,occlusion. & sec \\
        \hline \hline
        \multicolumn{4}{|l|}{\textit{\textbf{Reliability \& Stability}}} \\
        \hline
        INT & Initialization Time & Time from system start to first stable pose report. & sec \\
        \hline
        RLT & Re-acquisition Time (Loss) & Time to regain tracking after exiting and re-entering tracking \,\,volume. & sec \\
        \hline
        DRT & Long-Term Drift & Accumulated deviation at a static pose over extended periods \,\,\,(e.g., 30–60 minutes). & mm/min, deg/min \\
        \hline
    \end{tabularx}
\end{table}

Our evaluation methodology distinguishes between two crucial aspects of performance: absolute accuracy and relative precision. Absolute accuracy (e.g., AP) measures the overall end-to-end error from the ground truth, which is influenced by the one-time system calibration for each run. Relative precision, in contrast, quantifies the inherent consistency and stability of the tracking system itself, independent of this correctable calibration offset. We quantify this precision using Position Repeatability (RP) for static poses and the 3D 1-sigma trajectory error for dynamic paths. As both are calculated as the standard deviation of positional errors, they directly represent the system's stability, or scatter.

Both aspects are critical for a comprehensive assessment. While absolute accuracy reflects the practical performance in a given trial, this study places a particular emphasis on analyzing relative precision. This focus allows us to characterize the inherent performance of the controller's tracking algorithms, which cannot be corrected by the user.

A comprehensive set of metrics was used to assess accuracy, repeatability, robustness, and stability (see Table \ref{tab:evaluation_metrics}). Primary metrics include Position Accuracy (AP), Position Repeatability (RP), and Trajectory Accuracy (TP). Robustness was evaluated via occlusion recovery tests. However, long-term drift (DRT) and static jitter (STB) could not be assessed due to the controller's non-configurable five-second energy-saving mode.

Transformation matrices to align the ML2 and OptiTrack coordinate frames were computed using the corner points of an ArUco marker affixed to the test environment. All computational procedures are available in the associated Git repository \cite{tuhh_gitlab}, including a data cleaning pipeline with Z-Score, IQR, and kinematic filters \cite{Feis2015}; all parameters are detailed in the repository's documentation. Accuracy-related metrics are reported as 1-sigma values, reflecting the standard deviation.

\section{RESULTS}
This section presents the quantitative results of our study. To provide a comprehensive assessment, we report both absolute accuracy (e.g., RMS error), which offers a practical performance baseline for each trial, and relative precision (e.g., repeatability). However, our discussion deliberately emphasizes the relative precision metrics, as these more directly reflect the intrinsic stability and failure characteristics of the tracking system, independent of correctable, calibration-related offsets.

\subsection{Static Tracking Accuracy}
The evaluation of the ML2 controller's static tracking performance, based on the methodology described in Section \ref{Sec:Experimental Methodology} yielded the key results summarized in Table \ref{tab:accuracy_static}.

\begin{table*}[!htbp] 
\begin{threeparttable}
  \caption{Tracking accuracy and repeatability for each static pose (SP01–SP08) across all six experimental conditions (mm).}
  \label{tab:accuracy_static}
  \centering
\fontsize{6}{10}\selectfont 
\renewcommand{\arraystretch}{1.1}
\setlength{\tabcolsep}{3pt} 
  
 \setlength{\tabcolsep}{3pt} 
  \settowidth\rotheadsize{\parbox{1.8cm}{\centering Mean Acc.}} 
  \newcommand{\ROT}[2]{\rotatebox{#1}{\parbox{#2}{\centering}}}
  \begin{tabular}{@{} l *{18}{S[table-format=3.1, table-align-text-after=false]} *{3}{S[table-format=3.1, table-align-text-after=false]} @{}}
    \toprule
    \textbf{Pose ID} &
    \multicolumn{3}{c}{\textbf{F1 (25 mm/s)}} &
    \multicolumn{3}{c}{\textbf{F1 (50 mm/s)}} &
    \multicolumn{3}{c}{\textbf{F2 (25 mm/s)}} &
    \multicolumn{3}{c}{\textbf{F2 (50 mm/s)}} &
    \multicolumn{3}{c}{\textbf{S3 (25 mm/s)}} &
    \multicolumn{3}{c}{\textbf{S4 (25 mm/s)}} &
    \multicolumn{3}{c}{\textbf{Average (all conditions)}} \\ 
    \cmidrule(lr){2-4} \cmidrule(lr){5-7} \cmidrule(lr){8-10} \cmidrule(lr){11-13} \cmidrule(lr){14-16} \cmidrule(lr){17-19} \cmidrule(lr){20-22}
    & {\rotcell{\textbf{Mean Acc.}}} & {\rotcell{\textbf{Max Error}}} & {\rotcell{\textbf{Repeat.}}}
    & {\rotcell{\textbf{Mean Acc.}}} & {\rotcell{\textbf{Max Error}}} & {\rotcell{\textbf{Repeat.}}}
    & {\rotcell{\textbf{Mean Acc.}}} & {\rotcell{\textbf{Max Error}}} & {\rotcell{\textbf{Repeat.}}}
    & {\rotcell{\textbf{Mean Acc.}}} & {\rotcell{\textbf{Max Error}}} & {\rotcell{\textbf{Repeat.}}}
    & {\rotcell{\textbf{Mean Acc.}}} & {\rotcell{\textbf{Max Error}}} & {\rotcell{\textbf{Repeat.}}}
    & {\rotcell{\textbf{Mean Acc.}}} & {\rotcell{\textbf{Max Error}}} & {\rotcell{\textbf{Repeat.}}}
    & {\rotcell{\textbf{Mean Acc.}}} & {\rotcell{\textbf{Max Error}}} & {\rotcell{\textbf{Repeat.}}} \\ 
    \midrule
SP01 & 2.6 & 3.2 & 0.4 & 2.3 & 7.6 & 1.2 & 4.7 & 5.3 & 0.5 & 4.5 & 16.7 & 2.3 & 6.6 & 7.4 & 0.3 & 1.8 & 2.1 & 0.3 & 3.8 & 7.1 & 0.8 \\
SP02 & 3.6 & 4.0 & 0.4 & 1.0 & 6.7 & 1.1 & 2.1 & 2.5 & 0.6 & 0.5 & 1.0 & 0.4 & 4.5 & 4.6 & 0.2 & 2.4 & 3.1 & 0.3 & 2.4 & 3.7 & 0.5 \\
SP03 & 3.3 & 4.2 & 0.4 & 3.7 & 16.6 & 2.9 & 1.9 & 3.2 & 0.6 & 2.4 & 3.1 & 0.6 & 3.2 & 3.6 & 0.3 & 0.8 & 1.5 & 0.4 & 2.6 & 5.4 & 0.9 \\
SP04 & 1.9 & 2.6 & 0.9 & 1.6 & 8.3 & 1.1 & 2.8 & 10.7 & 1.0 & 1.3 & 1.7 & 0.5 & 5.5 & 6.0 & 0.6 & 2.3 & 2.6 & 0.5 & 2.6 & 5.3 & 0.8 \\
SP05 & 3.4 & 4.0 & 0.4 & 2.9 & 8.5 & 1.4 & 3.1 & 3.5 & 0.5 & 1.5 & 2.4 & 1.1 & 346.9* & 347.4* & 0.3 & 347.5* & 348.1* & 0.3 & 2.7 & 4.6 & 0.7 \\
SP06 & 4.3 & 12.2 & 3.0 & 2.3 & 8.1 & 1.1 & 2.6 & 4.5 & 1.5 & 1.3 & 1.8 & 0.5 & 346.8* & 347.6* & 0.4 & 346.5* & 347.5* & 0.4 & 2.6 & 6.6 & 1.2 \\
SP07 & 3.4 & 4.4 & 0.8 & 1.5 & 7.5 & 1.1 & 1.6 & 2.3 & 0.7 & 3.0 & 4.2 & 0.7 & 6.8 & 6.9 & 0.3 & 2.5 & 3.1 & 0.5 & 3.2 & 4.7 & 0.7 \\
SP08 & 4.5 & 5.0 & 0.9 & 2.6 & 7.5 & 1.3 & 3.2 & 4.0 & 0.6 & 2.6 & 3.7 & 0.5 & 4.0 & 4.3 & 0.2 & 2.0 & 2.4 & 0.2 & 3.0 & 4.5 & 0.7 \\
    \midrule 
    \textbf{Average} & 3.4 & 5.0 & 0.9 & 2.2 & 8.9 & 1.4 & 2.8 & 4.5 & 0.8 & 2.1 & 4.3 & 0.8 & 5.1 & 5.5 & 0.3 & 2.0 & 2.5 & 0.4 & 2.9 & 5.2 & 0.8 \\
    \bottomrule
  \end{tabular}
  \vspace{1ex}
  \raggedright
    \begin{tablenotes}
            \item[*] \footnotesize Note: The exceptionally large values for Mean Accuracy and Max Error were excluded from the calculation  as they represent outliers due to tracking limitations.
        \end{tablenotes}
    \end{threeparttable} 
\end{table*}

The static tracking evaluation revealed a clear dependency on the viewing angle and approach speed. Under optimal, frontal HMD positions, the system consistently achieved high accuracy. For instance, in the F2 condition at an approach speed of 50 mm/s, the mean error across all poses was only 2.1 mm. Higher approach speeds were generally associated with slight improvements in these optimal conditions.

Despite the generally robust performance, a critical failure mode observed under the angled viewing condition S3 significantly diminished its reliability. While most poses remained accurate, SP05 and SP06 exhibited gross positional errors, with a mean deviation of up to 347 mm. The most significant finding is that this major error was coupled with excellent repeatability (0.3-0.4 mm) (see Table \ref{tab:accuracy_static}). This indicates a systematic failure where the system confidently and consistently reports a severely incorrect position.

Finally, an analysis of the average performance across all conditions (with the major S3 outliers excluded, as noted in Table \ref{tab:accuracy_static}) reveals general trends. On average, poses SP01, SP07, and SP08 showed the highest mean errors. However, it is noteworthy that their performance was not substantially worse than that of the best-performing average pose (SP02 at 2.4 mm), indicating a generally consistent level of difficulty across the poses in non-failure conditions.

\subsection{Dynamic Tracking Accuracy}

This section presents the empirical findings from the tracking system's performance evaluation across 32 distinct experimental trials. The system's performance was quantified using several error metrics, including measurement precision (3D 1-sigma error), overall deviation (RMS error), maximum observed error, and temporal drift, all analyzed with respect to varying measurement positions, trajectory types, and movement speeds (see Table \ref{tab:accuracy_dynamic}).

\begin{table*}[htbp] 
\begin{threeparttable}
\caption{Summary of tracking metrics for 32 trials, grouped by trajectory class.}
\label{tab:accuracy_dynamic}
\centering
\fontsize{6}{10}\selectfont 
\renewcommand{\arraystretch}{1.1} 
\setlength{\tabcolsep}{3pt} 

\setlength{\tabcolsep}{2pt} 
\begin{tabular}{@{} p{0.6cm} p{1.1cm} p{1.0cm} p{1.1cm} *{4}{S[table-format=3.3]} *{4}{S[table-format=3.3]} *{4}{S[table-format=3.3]} *{1}{S[table-format=1.4]} @{}}
    \toprule
    \textbf{ID} & \textbf{Trajectory} & \textbf{Position} & \textbf{Speed [mm/s]} &
    \multicolumn{4}{c}{\textbf{Sigma 1 Error [mm]}} &
    \multicolumn{4}{c}{\textbf{RMS [mm]}} &
    \multicolumn{4}{c}{\textbf{Max Error [mm]}} &
    \textbf{3D-Drift [mm/s]} \\
    \cmidrule(lr){5-8} \cmidrule(lr){9-12} \cmidrule(lr){13-16} \cmidrule(lr){17-17} 
    & & & & {X} & {Y} & {Z} & {3D} &
    {X} & {Y} & {Z} & {3D} &
    {X} & {Y} & {Z} & {3D} &
    {3D} \\
    \midrule

T01 & circle & F1 & 10 & 1.610 & 2.620 & 1.460 & 3.040 & 2.080 & 2.620 & 3.000 & 4.500 & 3.860 & 3.650 & 4.410 & 5.750 & 0.0023 \\
T02 & circle & F2 & 10 & 0.920 & 0.510 & 1.440 & 1.530 & 1.650 & 1.460 & 4.160 & 4.710 & 2.620 & 2.150 & 6.130 & 6.640 & 0.0002 \\
T03 & circle & S3 & 10 & 1.170 & 0.430 & 0.480 & 1.010 & 1.860 & 0.880 & 0.500 & 2.120 & 2.920 & 1.340 & 0.750 & 3.000 & -0.0003 \\
T04 & circle & S4 & 10 & 1.640 & 0.860 & 0.570 & 1.590 & 3.200 & 2.600 & 0.760 & 4.190 & 4.730 & 3.420 & 1.250 & 5.920 & -0.0000 \\
T05 & circle & F1 & 50 & 0.500 & 0.580 & 1.990 & 1.170 & 1.500 & 2.710 & 3.040 & 4.340 & 2.530 & 4.990 & 8.060 & 8.650 & -0.0009 \\
T06 & circle & F2 & 50 & 0.970 & 0.830 & 1.120 & 1.010 & 1.190 & 0.840 & 2.720 & 3.080 & 1.690 & 1.360 & 4.090 & 4.600 & -0.0022 \\
T07 & circle & S4 & 50 & 0.870 & 0.450 & 0.410 & 0.340 & 0.910 & 0.870 & 1.060 & 1.650 & 1.490 & 1.430 & 1.830 & 2.370 & 0.0011 \\
\midrule
\multicolumn{4}{r}{\textbf{\textit{Average Circle}}} & \textit{1.100} & \textit{0.900} & \textit{1.067} & \textit{1.384} & \textit{1.770} & \textit{1.711} & \textit{2.177} & \textit{3.513} & \textit{2.834} & \textit{2.620} & \textit{3.789} & \textit{5.276} & \textit{0.0000} \\
\midrule

T08 & line & F1 & 10 & 0.330 & 0.450 & 0.500 & 0.380 & 5.120 & 1.040 & 1.100 & 5.340 & 6.070 & 2.470 & 2.460 & 6.470 & -0.0003 \\
T09 & line & F2 & 10 & 0.560 & 0.970 & 0.690 & 0.750 & 3.630 & 1.360 & 1.400 & 4.120 & 5.460 & 3.120 & 3.920 & 6.480 & 0.0003 \\
T10 & line & S3 & 10 & 0.670 & 0.400 & 0.140 & 0.380 & 0.730 & 0.480 & 0.140 & 0.880 & 2.070 & 1.840 & 0.460 & 2.100 & -0.0003 \\
T11 & line & S4 & 10 & 0.940 & 0.580 & 0.290 & 0.790 & 1.730 & 2.840 & 0.830 & 3.420 & 4.170 & 4.270 & 1.400 & 5.630 & 0.0016 \\
T12 & line & F1 & 50 & 0.440 & 0.590 & 0.650 & 0.580 & 0.660 & 0.610 & 0.910 & 1.280 & 1.470 & 1.630 & 2.220 & 2.650 & -0.0035 \\
T13 & line & F2 & 50 & 0.640 & 0.630 & 0.680 & 0.650 & 0.640 & 0.640 & 3.120 & 3.240 & 1.880 & 2.090 & 5.080 & 5.380 & -0.0054 \\
T14 & line & S3 & 50 & 0.760 & 0.500 & 0.150 & 0.290 & 0.920 & 0.500 & 1.160 & 1.560 & 1.490 & 1.540 & 2.760 & 2.940 & -0.0001 \\
T15 & line & S4 & 50 & 0.850 & 0.740 & 0.170 & 0.620 & 1.300 & 1.000 & 0.570 & 1.730 & 2.800 & 2.400 & 1.010 & 3.430 & 0.0162 \\
\midrule
\multicolumn{4}{r}{\textbf{\textit{Average Line}}} & \textit{0.649} & \textit{0.608} & \textit{0.409} & \textit{0.555} & \textit{1.841} & \textit{1.059} & \textit{1.154} & \textit{2.696} & \textit{3.176} & \textit{2.420} & \textit{2.414} & \textit{4.385} & \textit{0.0011} \\
\midrule
T16 & raster & F1 & 10 & 0.680 & 0.360 & 0.560 & 0.740 & 11.070 & 8.270 & 7.280 & 15.620 & 12.820 & 9.670 & 9.210 & 17.810 & 0.0012 \\
T17 & raster & F2 & 10 & 1.500 & 0.930 & 0.610 & 1.340 & 13.020 & 8.460 & 6.050 & 16.660 & 18.280 & 11.450 & 8.660 & 22.050 & 0.0001 \\
T21 & raster & F2 & 10* & 0.500 & 0.320 & 0.300 & 0.440 & 14.320 & 7.610 & 6.580 & 17.500 & 19.340 & 10.290 & 9.170 & 22.840 & 0.0000 \\
T18 & raster & S3 & 10 & 0.740 & 0.280 & 0.250 & 0.650 & 11.830 & 8.540 & 7.990 & 16.630 & 14.010 & 9.320 & 8.910 & 18.490 & -0.0006 \\
T19 & raster & S4 & 10 & 1.050 & 0.540 & 0.410 & 0.930 & 12.140 & 7.620 & 6.000 & 15.540 & 14.780 & 10.380 & 8.010 & 18.370 & -0.0003 \\
T20 & raster & F1 & 50 & 0.720 & 0.530 & 0.690 & 0.760 & 10.820 & 7.130 & 4.750 & 13.800 & 14.860 & 9.990 & 6.470 & 18.170 & -0.0018 \\
T22 & raster & F2 & 50 & 1.650 & 0.960 & 0.770 & 1.450 & 14.390 & 7.320 & 6.540 & 17.420 & 19.460 & 9.990 & 9.140 & 22.750 & 0.0032 \\
T23 & raster & S3 & 50 & 0.670 & 0.230 & 0.210 & 0.620 & 12.760 & 8.010 & 8.150 & 17.130 & 17.070 & 10.640 & 10.740 & 21.680 & 0.0024 \\
T24 & raster & S4 & 50 & 0.690 & 0.300 & 0.280 & 0.680 & 12.570 & 6.300 & 7.860 & 16.110 & 14.390 & 7.420 & 8.840 & 17.860 & -0.0001 \\
\midrule
\multicolumn{4}{r}{\textbf{\textit{Average Raster}}} & \textit{0.911} & \textit{0.494} & \textit{0.453} & \textit{0.846} & \textit{12.547} & \textit{7.696} & \textit{6.800} & \textit{16.268} & \textit{16.112} & \textit{9.894} & \textit{8.806} & \textit{20.002} & \textit{0.0004} \\
\midrule

T25 & square & F1 & 10 & 0.790 & 0.920 & 1.790 & 1.550 & 1.300 & 2.400 & 5.050 & 5.740 & 2.270 & 3.320 & 7.150 & 8.040 & 0.0002 \\
T26 & square & F2 & 10 & 1.200 & 1.180 & 1.790 & 1.150 & 2.280 & 1.910 & 2.620 & 3.970 & 3.550 & 3.120 & 4.700 & 6.010 & 0.0002 \\
T27 & square & S3 & 10 & 1.130 & 0.570 & 0.450 & 0.860 & 1.510 & 0.790 & 0.520 & 1.780 & 2.450 & 1.250 & 0.830 & 2.820 & 0.0002 \\
T28 & square & S4 & 10 & 0.660 & 1.270 & 1.010 & 0.800 & 0.900 & 1.690 & 1.100 & 2.210 & 1.440 & 2.850 & 1.930 & 3.510 & -0.0000 \\
T29 & square & F1 & 50 & 1.080 & 1.260 & 1.420 & 1.260 & 1.090 & 1.300 & 3.240 & 3.650 & 1.840 & 2.200 & 4.540 & 5.120 & 0.0003 \\
T30 & square & F2 & 50 & 1.160 & 1.090 & 1.100 & 1.090 & 1.160 & 1.120 & 3.830 & 4.160 & 3.510 & 3.250 & 7.060 & 7.220 & 0.0020 \\
T31 & square & S3 & 50 & 0.640 & 0.590 & 0.580 & 0.590 & 0.690 & 1.230 & 0.630 & 1.540 & 1.130 & 2.050 & 1.060 & 2.500 & 0.0002 \\
T32 & square & S4 & 50 & 0.700 & 1.090 & 0.950 & 0.690 & 0.800 & 1.100 & 1.400 & 1.960 & 1.280 & 1.760 & 2.050 & 2.790 & -0.0013 \\
\midrule
\multicolumn{4}{r}{\textbf{\textit{Average Square}}} & \textit{0.920} & \textit{0.996} & \textit{1.136} & \textit{0.999} & \textit{1.216} & \textit{1.443} & \textit{2.299} & \textit{3.126} & \textit{2.184} & \textit{2.475} & \textit{3.665} & \textit{4.751} & \textit{0.0002} \\
\bottomrule
\end{tabular}

    \begin{tablenotes}
            \item[*] \footnotesize Note: 50 Hz sampling rate used.
        \end{tablenotes}

    \end{threeparttable} 

\end{table*}

\begin{figure}
    \centering
    \includegraphics[width=1\linewidth]{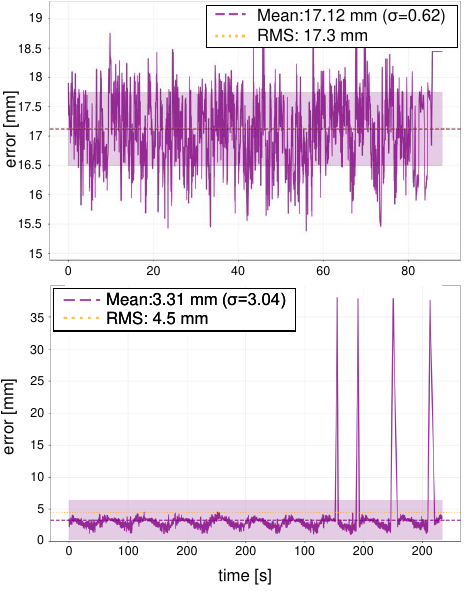}
    \caption{Top: Deviation of Raster-trajectory (S3;50 mm/s), Bottom: Deviation of Circle-trajectory (F1;10 mm/s).}
    \label{fig:dynamic_plot_example}
\end{figure}

\subsubsection{Accuracy Metrics:}

\textit{Absolute Accuracy (RMS Error):} Regarding absolute accuracy, the 3D RMS error quantifies the average path deviation. Line trajectories consistently performed best, showing the lowest average 3D RMS error (2.70 mm). In contrast, raster trajectories were the least accurate, with a significantly higher average RMS error of 16.27 mm. This is particularly noteworthy as the raster paths were executed in close proximity to the line trajectories, which suggests a substantial systematic deviation from the intended path that is specific to the raster motion pattern, despite the high precision of the path itself.

\textit{Sigma 1 Error (Precision/Scatter):} Use-case specific trajectories (line and raster) consistently showed the highest precision, with the lowest average 3D 1-sigma error (approx. 0.55 mm). In contrast, circle trajectories exhibited greater 3D 1-sigma error (approx. 1.38 mm), indicating reduced precision during continuous curved motion. Higher speeds (50 mm/s) were generally associated with slightly lower Sigma 1 errors compared to 10 mm/s, suggesting improved stability during dynamic tracking (see Figure \ref{fig:dynamic_plot_example}). Spatially, precision was best at closer distances, with S3 showing the lowest average 3D 1-sigma error (approx. 0.81 mm), while the 600 mm (front) position had the highest (approx. 1.05 mm). The most precise result was a 3D Sigma 1 error of 0.290 mm (S3, line trajectory, 50 mm/s).

\textit{Error Distribution and Maximum Deviation:} To analyze the nature of the errors, we also considered the maximum deviation. This analysis revealed significant peaks that persisted even after 3σ outlier removal, particularly in raster trajectories. These trajectories consistently showed the largest delta between the RMS and 1-sigma error, suggesting the high overall RMS is driven by a few, high-magnitude outliers. This trend aligned with the RMS results, with the 600 mm (front) raster trajectory showing the highest errors. The absolute maximum 3D error identified in the raw data, prior to filtering, was 34.72 mm during a circular trajectory (see Figure \ref{fig:dynamic_plot_example} and \cite{tuhh_gitlab}).

\subsubsection{Drift (Temporal Stability):} 
The positional drift, indicating the change in measured position over time, was found to be minimal across most experimental conditions, with average 3D drift rates approaching zero. This suggests a high degree of temporal stability in the system's tracking performance over the duration of the trials.

\subsubsection{Angular Tracking Performance:}
In addition to positional metrics, comprehensive angular tracking data (orientation errors in roll, pitch, and yaw) was also collected during the trials. Due to space constraints within this publication, the detailed angular error data is not presented here but is fully available in the accompanying Git repository. It is important to note that the observed angular error was consistently and significantly below our predefined operational threshold of 10 degrees, indicating robust and accurate orientation tracking across all tested conditions.

\subsubsection{Robustness, Reliability, Stability}

\begin{figure}
    \centering
    \includegraphics[width=1\linewidth]{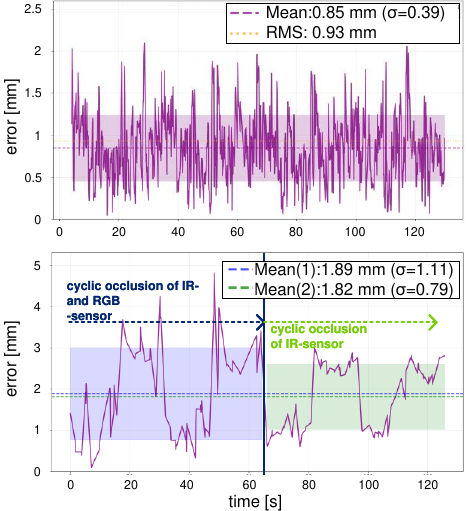}
    \caption{Top: Plot of a line trajectory (P4; 10 mm/s) without disturbance; Bottom: Occlusion disturbance on a line trajectory (P4; 10 mm/s)}
    \label{fig:occlusion}
\end{figure}

The system's robustness to occlusion was found to be highly velocity-dependent. At the lower speed of 10 mm/s, the system demonstrated a 100\% Occlusion Success Rate (OSR), re-acquiring the track after a 5-second occlusion with a negligible Re-acquisition Time (ROT). During these 10 mm/s tests, a slight degradation in precision was noted. The tracking remained highly precise when only the IR sensors were occluded (3D 1-sigma error of 0.79 mm), with a more pronounced, yet still robust, decrease in precision when both camera and IR sensors were blocked (3D 1-sigma error of 1.11 mm). At the higher speed of 50 mm/s, however, no stable tracking could be recovered, resulting in an OSR of 0

The Re-acquisition Time after tracking loss (RLT) was consistently under one second. While a precise quantification was challenging due to the undefined boundaries of the tracking volume, this test revealed a key system strength: even during these re-acquisition events, the system maintained a high precision, with a measured 3D 1-sigma error of only 1.52 mm.

Regarding system stability, the Initialization Time (INT) was consistently below one second until a valid pose was reported. As noted in section \ref{Metrics and Data Evaluation}, long-term static drift (DRT) could not be measured due to technical constraints.

\section{DISCUSSION}
The results provide a comprehensive performance baseline for the ML2 controller. This section interprets these findings by comparing them to the state of the art, discussing the implications for our industrial use case, and outlining the study's limitations.

\subsection{Interpretation of Key Findings}

The results confirm that while the ML2 controller can achieve high tracking accuracy under optimal conditions, its performance is highly dependent on specific factors, revealing critical failure modes.

The most significant finding is a systematic failure at specific static poses under angled viewing conditions. The system exhibited gross positional errors despite high precision. This suggests that the failure is not due to random sensor noise (jitter), but is a systematic failure. We hypothesize this is caused by insufficient environmental features for the controller's SLAM system, leading to a stable but incorrect pose. This represents a critical failure mode, as a confident but erroneous signal is more dangerous for user guidance than observable jitter.

In dynamic tests, performance was also conditional. Simpler trajectories (line, raster) yielded higher accuracy than more complex paths (circle, square), and higher speeds counter-intuitively improved precision. It is plausible that at very low speeds, the tracking estimate is more susceptible to sensor noise. This suggests the system's sensor fusion algorithm more heavily weights IMU data during continuous motion, effectively filtering sensor noise. While the system demonstrated robust re-acquisition after tracking loss at low speeds, maintaining a clear line of sight proved critical for preventing tracking failures at higher velocities.

\subsection{Comparison with the State of the Art}
The achieved mean dynamic accuracy in the \textless 2 mm confirms preliminary investigations \cite{Masuhr2024} and positions the markerless ML2 controller favorably against general-purpose hand tracking, which often exhibits errors exceeding 20\ mm \cite{Soares2021HL2HandTracking}. The results of the initial study for static poses could be confirmed \cite{Masuhr2024}. However, it does not reach the sub-millimeter precision reported for specialized, marker-based inside-out systems developed for surgical applications \cite{Gsaxner2021, MartinGomez2024STTAR}. This study therefore quantifies a key trade-off: a potential sacrifice of 1-2\,mm in absolute accuracy is exchanged for the significant operational advantages of a fully integrated, commercially available system with minimal setup time. 

\subsection{Implications for the Industrial Use Case}
Since the experimental environment and parameters were designed to resemble those of a leak testing scenario, the results can be meaningfully transferred to the intended application. For the hydrogen leak inspection task, the controller's performance under optimal, frontal viewing conditions consistently yielded errors well below the predefined 5\,mm requirement. However, this level of accuracy was not maintained across all configurations, with significant positional outliers occurring under specific non-frontal viewing angles. Based on these results, the ML2 controller cannot be considered reliable for this safety-critical task. A practical implementation would require a fail-safe mechanism capable of detecting this specific, high-precision failure mode, which goes beyond simple outlier filtering.

Further practical challenges arise from using an off-the-shelf industrial solution. The non-configurable energy-saving mode, which disables tracking after five seconds of inactivity, hinders workflows that involve pauses. This highlights the trade-off between the low setup effort of a commercial system and the lack of customizability for specific research or process needs. The controller maintained robust tracking performance despite the 3D-printed mechanical attachment, essential for leakage inspection and requiring a front-mounted gas inlet. Like the leakage inspection, other use-cases would also require dedicated attachments. This validates the practical feasibility of an integrated inspection tool, though a production-level design needs improved robustness and compactness.

Crucially, the system's inherent tracking precision can be evaluated independently of its absolute calibration. While our results show high precision (low scatter), the systematic RMS error requires a separate, user-performed calibration step to be eliminated. The performance of this calibration is a separate research question that does not alter our findings on the controller's tracking stability.

\subsection{Limitations}
This study was conducted in a controlled laboratory environment. Performance in real industrial settings may differ, where factors such as dust, vibrations, and variable lighting conditions are present. Furthermore, the evaluation was limited to a single ML2 device, which may restrict the generalizability of the results.

Several methodological choices further constrain the findings. To reduce experimental complexity, the controller’s orientation was kept constant throughout most of the tests. A comprehensive 6-DoF error analysis under varying orientations remains the subject of future work. Additionally, the re-calibration process using an ArUco marker, although effective, introduces a necessary manual step each time the HMD position is changed.

A further methodological consideration is the asynchronous data streams (ML2 at \textasciitilde 30 Hz, OptiTrack at 50 Hz), which we aligned using direct timestamp-based correlation instead of temporal interpolation. We validated this approach by calculating the maximum theoretical position error from the temporal discrepancy (0.25 mm at 50 mm/s). As this value lies within the OptiTrack's own accuracy range (\textasciitilde 0.2 mm), its influence is considered negligible. However, the effect of this temporal uncertainty appears to be more pronounced at lower sampling rates, potentially indicating a measurement artifact. This was evident when an identical slow trajectory yielded a 3D 1-sigma error of 1.34 mm when measured at 10 Hz, versus 0.44 mm at 50 Hz (see Table \ref{tab:accuracy_dynamic}). This discrepancy indicates that the 50 Hz data more accurately represents the system's true precision.

\section{CONCLUSION AND FUTURE WORK}

In conclusion, this work presents the first comprehensive benchmark of the ML2 controller for industrial tool tracking, applying a robot-driven, standards-compliant methodology. The results demonstrate that the controller is a capable off-the-shelf solution, achieving accuracy of \textless 2\,mm under ideal conditions, sufficient for critical tasks such as hydrogen leak inspection. However, the system’s reliability is highly dependent on the viewing angle and controller pose. This study quantifies a key trade-off between the operational flexibility of a fully integrated, markerless system and the maximum achievable precision of more constrained tracking solutions. Crucially, because these severe and systematic failure modes are detectable, they can be addressed through software-level mitigation.

Based on these findings and the identified limitations, several key aspects for future work emerge. Further research should focus on improving core tracking performance. This involves two key directions: investigating the root causes of the observed systematic errors, and exploring real-time post-processing to smooth high-frequency jitter with filters (e.g., Kalman) and mitigate gross errors via robust outlier detection. Methodologically, the evaluation should be expanded to include a full 6-DoF analysis under varying tool orientations. Most critically, bridging the gap to real-world application requires conducting user studies to assess the impact of the measured accuracy on task performance. More robust mechanical and software solutions must be developed to address the practical limitations of the commercial hardware, such as its non-configurable energy-saving mode. To ensure the necessary end-to-end accuracy, it's crucial to examine the registration accuracy, for example, with typical components and conditions relevant to the use case. Future work should also assess the calibration performance itself. This could involve evaluating user-centric methods, such as touching physical reference points, to quantify the achievable end-to-end accuracy in practice.

\bibliographystyle{abbrv-doi}

\bibliography{bibliography}

\begin{thebibliography}{10}

\bibitem{Akpasi2025}
S.~O. Akpasi, I.~M.~S. Anekwe, E.~K. Tetteh, U.~O. Amune, S.~I. Mustapha, and S.~L. Kiambi.
\newblock Hydrogen as a clean energy carrier: advancements, challenges, and its role in a sustainable energy future.
\newblock {\em Clean Energy}, 9(1):52--88, 2025. doi: {{%
10\hspace{.1pt}\discretionary{.}{%
}{.}\hspace{.4pt}1093\discretionary{/}{%
}{/}ce\discretionary{/}{%
}{/}zkae112}}


\bibitem{Azuma1997}
R.~T. Azuma.
\newblock A survey of augmented reality.
\newblock {\em Presence: Teleoperators and Virtual Environments}, 6(4):355--385, 1997. doi: {{%
10\hspace{.1pt}\discretionary{.}{%
}{.}\hspace{.4pt}1162\discretionary{/}{%
}{/}pres\hspace{.1pt}\discretionary{.}{%
}{.}\hspace{.4pt}1997\hspace{.1pt}\discretionary{.}{%
}{.}\hspace{.4pt}6\hspace{.1pt}\discretionary{.}{%
}{.}\hspace{.4pt}4\hspace{.1pt}\discretionary{.}{%
}{.}\hspace{.4pt}355}}


\bibitem{Bertovic2015}
M.~Bertovic.
\newblock {\em Human Factors in Non-Destructive Testing (NDT): Risks and Challenges of Mechanised NDT}.
\newblock Federal Institute For Materials Research and Testing, Berlin, 2015. doi: {{%
10\hspace{.1pt}\discretionary{.}{%
}{.}\hspace{.4pt}14279\discretionary{/}{%
}{/}depositonce\discretionary{%
}{-}{-}4685}}


\bibitem{Boal2023}
M.~Boal, D.~Anastasiou, F.~Tesfai, W.~Ghamrawi, E.~Mazomenos, N.~Curtis, J.~Collins, A.~Sridhar, J.~Kelly, D.~Stoyanov, and N.~Francis.
\newblock Evaluation of objective tools and artificial intelligence in robotic surgery technical skills assessment: a systematic review.
\newblock {\em The British Journal of Surgery}, 111(1), 2023. doi: {{%
10\hspace{.1pt}\discretionary{.}{%
}{.}\hspace{.4pt}1093\discretionary{/}{%
}{/}bjs\discretionary{/}{%
}{/}znad331}}


\bibitem{Brand2020}
M.~Brand, L.~A. Wulff, Y.~Hamdani, and T.~Schüppstuhl.
\newblock Accuracy of marker tracking on an optical see-through head mounted display.
\newblock In T.~Schüppstuhl, K.~Tracht, and D.~Henrich, eds., {\em Annals of Scientific Society for Assembly, Handling and Industrial Robotics}, pp. 19--28. Springer Vieweg, Berlin, Heidelberg, 2020. doi: {{%
10\hspace{.1pt}\discretionary{.}{%
}{.}\hspace{.4pt}1007\discretionary{/}{%
}{/}978\discretionary{%
}{-}{-}3\discretionary{%
}{-}{-}662\discretionary{%
}{-}{-}61755\discretionary{%
}{-}{-}7\_3}}


\bibitem{Campos2021ORBSLAM3}
C.~Campos, R.~Elvira, J.~J. Garc{\'{i}}a~Rodr{\'{i}}guez, J.~M.~M. Montiel, and J.~D. Tard{\'{o}}s.
\newblock {ORB-SLAM3}: An accurate open-source library for visual, visual-inertial and multi-map {SLAM}.
\newblock {\em IEEE Transactions on Robotics}, 37(6):1874--1890, 2021. doi: {{%
10\hspace{.1pt}\discretionary{.}{%
}{.}\hspace{.4pt}1109\discretionary{/}{%
}{/}TRO\hspace{.1pt}\discretionary{.}{%
}{.}\hspace{.4pt}2021\hspace{.1pt}\discretionary{.}{%
}{.}\hspace{.4pt}3075644}}


\bibitem{tuhh_gitlab}
{Christian Masuhr}.
\newblock {GIT Repository for ML2 accuracy study}.
\newblock Online; \url{https://collaborating.tuhh.de/ccm6674/vrst_2025_ml2_accuracy_study}, 2025.

\bibitem{Costa2024}
G.~M. Costa, M.~R. Petry, J.~G. Martins, and A.~P. G.~M. Moreira.
\newblock Assessment of multiple fiducial marker trackers on hololens 2.
\newblock {\em IEEE Access}, 12:14211--14226, 2024. doi: {{%
10\hspace{.1pt}\discretionary{.}{%
}{.}\hspace{.4pt}1109\discretionary{/}{%
}{/}ACCESS\hspace{.1pt}\discretionary{.}{%
}{.}\hspace{.4pt}2024\hspace{.1pt}\discretionary{.}{%
}{.}\hspace{.4pt}3356722}}


\bibitem{Passos2020}
D.~Eger~Passos and B.~Jung.
\newblock Measuring the accuracy of inside-out tracking in xr devices using a high-precision robotic arm.
\newblock In C.~Stephanidis and M.~Antona, eds., {\em HCI International 2020 - Posters}, pp. 19--26. Springer International Publishing, Cham, 2020. doi: {{%
10\hspace{.1pt}\discretionary{.}{%
}{.}\hspace{.4pt}1007\discretionary{/}{%
}{/}978\discretionary{%
}{-}{-}3\discretionary{%
}{-}{-}030\discretionary{%
}{-}{-}50726\discretionary{%
}{-}{-}8\_3}}


\bibitem{en_iso_9283}
{European Committee for Standardization/International Organization for Standardization}.
\newblock {EN ISO 9283: Manipulating industrial robots - Performance criteria and related test methods (ISO 9283:1998)}.
\newblock Standard EN ISO 9283, CEN/ISO, 1998.

\bibitem{Feis2015}
S.~Feis and G.~Gatti.
\newblock A simple method to choose the most representative stride and detect outliers.
\newblock {\em Computer Methods in Biomechanics and Biomedical Engineering}, 18(3):348--353, 2015. doi: {{%
10\hspace{.1pt}\discretionary{.}{%
}{.}\hspace{.4pt}1080\discretionary{/}{%
}{/}10255842\hspace{.1pt}\discretionary{.}{%
}{.}\hspace{.4pt}2013\hspace{.1pt}\discretionary{.}{%
}{.}\hspace{.4pt}801878}}


\bibitem{Frandsen2022}
J.~Frandsen, J.~Tenny, W.~Frandsen, R.~Thuesen, and R.~Kristensen.
\newblock An augmented reality maintenance assistant with real-time quality inspection on handheld mobile devices.
\newblock {\em International Journal of Advanced Manufacturing Technology}, 125(7-8):4253--4270, 2023. doi: {{%
10\hspace{.1pt}\discretionary{.}{%
}{.}\hspace{.4pt}1007\discretionary{/}{%
}{/}s00170\discretionary{%
}{-}{-}023\discretionary{%
}{-}{-}10978\discretionary{%
}{-}{-}4}}


\bibitem{GarridoJurado2014}
S.~Garrido-Jurado, R.~Mu{\~n}oz-Salinas, F.~J. Madrid-Cuevas, and M.~J. Mar{\'\i}n-Jim{\'e}nez.
\newblock Automatic generation and detection of highly reliable fiducial markers under occlusion.
\newblock {\em Pattern Recognition}, 47(6):2280--2292, 2014. doi: {{%
10\hspace{.1pt}\discretionary{.}{%
}{.}\hspace{.4pt}1016\discretionary{/}{%
}{/}j\hspace{.1pt}\discretionary{.}{%
}{.}\hspace{.4pt}patcog\hspace{.1pt}\discretionary{.}{%
}{.}\hspace{.4pt}2014\hspace{.1pt}\discretionary{.}{%
}{.}\hspace{.4pt}01\hspace{.1pt}\discretionary{.}{%
}{.}\hspace{.4pt}005}}


\bibitem{GaitBalance2024}
D.~J. Geerse, P.~F. van Doorn, J.~S. van Bergem, E.~M. Hoogendoorn, E.~Nyman, and M.~Roerdink.
\newblock Gait and balance assessments with augmented reality glasses in people with parkinson's disease: Concurrent validity and test–retest reliability.
\newblock {\em PLoS ONE}, 19(9):e0300976, 2024. doi: {{%
10\hspace{.1pt}\discretionary{.}{%
}{.}\hspace{.4pt}1371\discretionary{/}{%
}{/}journal\hspace{.1pt}\discretionary{.}{%
}{.}\hspace{.4pt}pone\hspace{.1pt}\discretionary{.}{%
}{.}\hspace{.4pt}0300976}}


\bibitem{Gsaxner2021}
C.~Gsaxner, J.~Thome, J.~Egger, D.~{\v{S}}tern, D.~Schmalstieg, and K.-M. Reichl.
\newblock Inside-out instrument tracking for surgical navigation in augmented reality.
\newblock {\em International Journal of Computer Assisted Radiology and Surgery}, 16(9):1475--1484, 2021. doi: {{%
10\hspace{.1pt}\discretionary{.}{%
}{.}\hspace{.4pt}1007\discretionary{/}{%
}{/}s11548\discretionary{%
}{-}{-}021\discretionary{%
}{-}{-}02422\discretionary{%
}{-}{-}x}}


\bibitem{Hu2022}
X.~Hu, F.~R.~y. Baena, and F.~Cutolo.
\newblock Head-mounted augmented reality platform for markerless orthopaedic navigation.
\newblock {\em IEEE Journal of Biomedical and Health Informatics}, 26(2):910--921, 2022. doi: {{%
10\hspace{.1pt}\discretionary{.}{%
}{.}\hspace{.4pt}1109\discretionary{/}{%
}{/}JBHI\hspace{.1pt}\discretionary{.}{%
}{.}\hspace{.4pt}2021\hspace{.1pt}\discretionary{.}{%
}{.}\hspace{.4pt}3088442}}


\bibitem{Inficion.RoboticSniffing}
{INFICON GmbH}.
\newblock White paper: Robotic sniffing.
\newblock 2019. url: https://www.inficon.com/en/service-and-support/whitepapers-and-e-books/robotic-leak-testing.

\bibitem{ISO18305}
{International Organization for Standardization}.
\newblock Iso/iec 18305:2016: Information technology — real-time locating systems — test and evaluation of localization and tracking systems.
\newblock Standard ISO/IEC 18305:2016, ISO/IEC, 2016.

\bibitem{Koch2024}
J.~Koch, D.~Jevremovic, K.~Moenck, and T.~Schüppstuhl.
\newblock A digital assistance system leveraging vision foundation models \& 3d localization for reproducible defect segmentation in visual inspection.
\newblock {\em Procedia CIRP}, 130:387--397, 2024. doi: {{%
10\hspace{.1pt}\discretionary{.}{%
}{.}\hspace{.4pt}1016\discretionary{/}{%
}{/}j\hspace{.1pt}\discretionary{.}{%
}{.}\hspace{.4pt}procir\hspace{.1pt}\discretionary{.}{%
}{.}\hspace{.4pt}2024\hspace{.1pt}\discretionary{.}{%
}{.}\hspace{.4pt}10\hspace{.1pt}\discretionary{.}{%
}{.}\hspace{.4pt}105}}


\bibitem{Koch2023}
J.~Koch, G.~Lotzing, H.~Eschen, M.~Lütjen, and T.~Schüppstuhl.
\newblock A human-centered hot platform approach for manual inspections.
\newblock {\em Procedia CIRP}, 119:840--845, 2023. doi: {{%
10\hspace{.1pt}\discretionary{.}{%
}{.}\hspace{.4pt}1016\discretionary{/}{%
}{/}j\hspace{.1pt}\discretionary{.}{%
}{.}\hspace{.4pt}procir\hspace{.1pt}\discretionary{.}{%
}{.}\hspace{.4pt}2023\hspace{.1pt}\discretionary{.}{%
}{.}\hspace{.4pt}09\hspace{.1pt}\discretionary{.}{%
}{.}\hspace{.4pt}072}}


\bibitem{Luckett2019}
E.~Luckett, T.~Key, N.~Newsome, and J.~A. Jones.
\newblock Metrics for the evaluation of tracking systems for virtual environments.
\newblock In {\em 2019 IEEE Conference on Virtual Reality and 3D User Interfaces (VR)}, pp. 1084--1085, 2019. doi: {{%
10\hspace{.1pt}\discretionary{.}{%
}{.}\hspace{.4pt}1109\discretionary{/}{%
}{/}VR\hspace{.1pt}\discretionary{.}{%
}{.}\hspace{.4pt}2019\hspace{.1pt}\discretionary{.}{%
}{.}\hspace{.4pt}8798374}}


\bibitem{Marino2021}
E.~Marino, L.~Barbieri, B.~Colacino, F.~G.~H. Ferreira, and A.~M.~T. Loureiro.
\newblock An ar inspection tool to support workers in i4.0 environments.
\newblock {\em Computers in Industry}, 128:103412, 2021. doi: {{%
10\hspace{.1pt}\discretionary{.}{%
}{.}\hspace{.4pt}1016\discretionary{/}{%
}{/}j\hspace{.1pt}\discretionary{.}{%
}{.}\hspace{.4pt}compind\hspace{.1pt}\discretionary{.}{%
}{.}\hspace{.4pt}2021\hspace{.1pt}\discretionary{.}{%
}{.}\hspace{.4pt}103412}}


\bibitem{Marino2022Benchmarking}
E.~Marino, F.~Bruno, L.~Barbieri, and A.~Lagudi.
\newblock Benchmarking built-in tracking systems for indoor ar applications on popular mobile devices.
\newblock {\em Sensors}, 22(14):5382, 2022. doi: {{%
10\hspace{.1pt}\discretionary{.}{%
}{.}\hspace{.4pt}3390\discretionary{/}{%
}{/}s22145382}}


\bibitem{MartinGomez2024STTAR}
A.~Martin-Gomez, H.~Li, T.~Song, S.~Yang, G.~Wang, H.~Ding, N.~Navab, Z.~Zhao, and M.~Armand.
\newblock {STTAR: Surgical Tool Tracking Using Off-the-Shelf Augmented Reality Head-Mounted Displays}.
\newblock {\em IEEE Transactions on Visualization and Computer Graphics}, 30(7):3578--3593, jul 2024.
\newblock Epub 2024 Jun 27. doi: {{%
10\hspace{.1pt}\discretionary{.}{%
}{.}\hspace{.4pt}1109\discretionary{/}{%
}{/}TVCG\hspace{.1pt}\discretionary{.}{%
}{.}\hspace{.4pt}2023\hspace{.1pt}\discretionary{.}{%
}{.}\hspace{.4pt}3238309}}


\bibitem{Masuhr2024}
C.~Masuhr, L.~Büsch, and T.~Schüppstuhl.
\newblock Leakage inspection for the scale-up of hydrogen electrolyzers: A case study and comparative analysis of technologies.
\newblock In F.~J.~G. Silva, A.~B. Pereira, and R.~D. S.~G. Campilho, eds., {\em Flexible Automation and Intelligent Manufacturing: Establishing Bridges for More Sustainable Manufacturing Systems. FAIM 2023}, Lecture Notes in Mechanical Engineering, pp. 661--672. Springer, Cham, 2024. doi: {{%
10\hspace{.1pt}\discretionary{.}{%
}{.}\hspace{.4pt}1007\discretionary{/}{%
}{/}978\discretionary{%
}{-}{-}3\discretionary{%
}{-}{-}031\discretionary{%
}{-}{-}38241\discretionary{%
}{-}{-}3\_56}}


\bibitem{MasuhrUsability}
C.~Masuhr, J.~Koch, and T.~Sch{\"u}ppstuhl.
\newblock Augmented reality authoring for efficient inspections in green aviation: Evaluating accuracy and usability.
\newblock In H.~Kohl, G.~Seliger, F.~Dietrich, and H.~T. Vien, eds., {\em Decarbonizing Value Chains}, pp. 321--329. Springer Nature Switzerland, Cham, 2025. doi: {{%
10\hspace{.1pt}\discretionary{.}{%
}{.}\hspace{.4pt}1007\discretionary{/}{%
}{/}978\discretionary{%
}{-}{-}3\discretionary{%
}{-}{-}031\discretionary{%
}{-}{-}93891\discretionary{%
}{-}{-}7\_36}}


\bibitem{Nancel2021HL2ViveComp}
M.~Nancel, B.~Ens, and J.~D. Hincapié-Ramos.
\newblock Comparing the htc vive, microsoft hololens and optitrack: Accuracy and precision in a large projective space.
\newblock In {\em Extended Abstracts of the 2021 CHI Conference on Human Factors in Computing Systems (CHI EA '21)}, pp. 1--7. ACM, New York, NY, USA, may 2021. doi: {{%
10\hspace{.1pt}\discretionary{.}{%
}{.}\hspace{.4pt}1145\discretionary{/}{%
}{/}3411763\hspace{.1pt}\discretionary{.}{%
}{.}\hspace{.4pt}3451612}}


\bibitem{polvi2018}
J.~Polvi, T.~Taketomi, A.~Moteki, T.~Yoshitake, T.~Fukuoka, G.~Yamamoto, C.~Sandor, and H.~Kato.
\newblock Handheld guides in inspection tasks: Augmented reality versus picture.
\newblock {\em IEEE Transactions on Visualization and Computer Graphics}, 24(7):2118--2128, 2018. doi: {{%
10\hspace{.1pt}\discretionary{.}{%
}{.}\hspace{.4pt}1109\discretionary{/}{%
}{/}TVCG\hspace{.1pt}\discretionary{.}{%
}{.}\hspace{.4pt}2017\hspace{.1pt}\discretionary{.}{%
}{.}\hspace{.4pt}2709746}}


\bibitem{Sa2022}
I.~S{\'a}, P.~Pimentel, G.~Guedes, P.~V.~K. Borges, and G.~N. DeSouza.
\newblock Benchmarking built-in tracking systems for indoor ar applications on popular mobile devices.
\newblock {\em Sensors}, 22(14):5331, 2022. doi: {{%
10\hspace{.1pt}\discretionary{.}{%
}{.}\hspace{.4pt}3390\discretionary{/}{%
}{/}s22145331}}


\bibitem{Sitole2020}
S.~P. Sitole, A.~K. LaPre, and F.~C. Sup.
\newblock Application and evaluation of lighthouse technology for precision motion capture.
\newblock {\em IEEE Sensors Journal}, 20(15):8576--8585, 2020. doi: {{%
10\hspace{.1pt}\discretionary{.}{%
}{.}\hspace{.4pt}1109\discretionary{/}{%
}{/}JSEN\hspace{.1pt}\discretionary{.}{%
}{.}\hspace{.4pt}2020\hspace{.1pt}\discretionary{.}{%
}{.}\hspace{.4pt}2983933}}


\bibitem{Soares2021HL2HandTracking}
J.~Soares, D.~Gessl, C.~Gstoettner, L.~A. Hruby, H.~Kiss, J.~Kleesiek, A.~R. Carnièl, and B.~Menze.
\newblock Accuracy and repeatability of microsoft hololens 2 for manual measurement tasks.
\newblock {\em Sensors}, 21(11):3743, may 2021. doi: {{%
10\hspace{.1pt}\discretionary{.}{%
}{.}\hspace{.4pt}3390\discretionary{/}{%
}{/}s21113743}}


\bibitem{Son2022}
Y.~Son, J.~Yeom, S.~Lim, D.-H. Kim, and K.-S. Choi.
\newblock Method and system for evaluating tracking performance of vr/ar/mr devices.
\newblock In {\em 2022 13th International Conference on Information and Communication Technology Convergence (ICTC)}, pp. 2074--2076, 2022. doi: {{%
10\hspace{.1pt}\discretionary{.}{%
}{.}\hspace{.4pt}1109\discretionary{/}{%
}{/}ICTC55196\hspace{.1pt}\discretionary{.}{%
}{.}\hspace{.4pt}2022\hspace{.1pt}\discretionary{.}{%
}{.}\hspace{.4pt}9952887}}


\bibitem{Sturm2012}
J.~Sturm, N.~Engelhard, F.~Endres, W.~Burgard, and D.~Cremers.
\newblock A benchmark for the evaluation of {RGB-D} {SLAM} systems.
\newblock In {\em Proceedings of the {IEEE/RSJ} International Conference on Intelligent Robots and Systems ({IROS})}, pp. 573--580, 2012. doi: {{%
10\hspace{.1pt}\discretionary{.}{%
}{.}\hspace{.4pt}1109\discretionary{/}{%
}{/}IROS\hspace{.1pt}\discretionary{.}{%
}{.}\hspace{.4pt}2012\hspace{.1pt}\discretionary{.}{%
}{.}\hspace{.4pt}6385773}}


\bibitem{TCHOUVELEV202112439}
A.~V. Tchouvelev, W.~J. Buttner, D.~Melideo, D.~Baraldi, and B.~Angers.
\newblock Development of risk mitigation guidance for sensor placement inside mechanically ventilated enclosures – phase 1.
\newblock {\em International Journal of Hydrogen Energy}, 46(23):12439--12454, 2021.
\newblock ICHS 2019 Conference. doi: {{%
10\hspace{.1pt}\discretionary{.}{%
}{.}\hspace{.4pt}1016\discretionary{/}{%
}{/}j\hspace{.1pt}\discretionary{.}{%
}{.}\hspace{.4pt}ijhydene\hspace{.1pt}\discretionary{.}{%
}{.}\hspace{.4pt}2020\hspace{.1pt}\discretionary{.}{%
}{.}\hspace{.4pt}09\hspace{.1pt}\discretionary{.}{%
}{.}\hspace{.4pt}108}}


\bibitem{Torres2021}
Y.~Torres, S.~Nadeau, and K.~Landau.
\newblock Classification and quantification of human error in manufacturing: A case study in complex manual assembly.
\newblock {\em Applied Sciences}, 11(2):749, 2021. doi: {{%
10\hspace{.1pt}\discretionary{.}{%
}{.}\hspace{.4pt}3390\discretionary{/}{%
}{/}app11020749}}


\bibitem{Mai2022}
M.~Trinh, N.~V. Navkar, and Z.~Deng.
\newblock A practical ar-based surgical navigation system using optical see-through head mounted display.
\newblock In {\em 2022 IEEE 22nd International Conference on Bioinformatics and Bioengineering (BIBE)}, pp. 164--167, 2022. doi: {{%
10\hspace{.1pt}\discretionary{.}{%
}{.}\hspace{.4pt}1109\discretionary{/}{%
}{/}BIBE55377\hspace{.1pt}\discretionary{.}{%
}{.}\hspace{.4pt}2022\hspace{.1pt}\discretionary{.}{%
}{.}\hspace{.4pt}00043}}


\bibitem{VanDoorn2025}
P.~F. van Doorn, D.~J. Geerse, J.~S. van Bergem, E.~M. Hoogendoorn, E.~Nyman, and M.~Roerdink.
\newblock Gait parameters can be derived reliably and validly from augmented reality glasses in people with parkinson's disease performing 10-m walk tests at comfortable and fast speeds.
\newblock {\em Sensors}, 25(4):1230, 2025. doi: {{%
10\hspace{.1pt}\discretionary{.}{%
}{.}\hspace{.4pt}3390\discretionary{/}{%
}{/}s25041230}}


\bibitem{Voinea2023}
G.-D. Voinea, F.~Gîrbacia, M.~Duguleană, R.~G. Boboc, and C.~Gheorghe.
\newblock Mapping the emergent trends in industrial augmented reality.
\newblock {\em Electronics}, 12(7):1719, 2023. doi: {{%
10\hspace{.1pt}\discretionary{.}{%
}{.}\hspace{.4pt}3390\discretionary{/}{%
}{/}electronics12071719}}


\bibitem{WANG2016406}
X.~Wang, S.~Ong, and A.~Nee.
\newblock Multi-modal augmented-reality assembly guidance based on bare-hand interface.
\newblock {\em Advanced Engineering Informatics}, 30(3):406--421, 2016. doi: {{%
10\hspace{.1pt}\discretionary{.}{%
}{.}\hspace{.4pt}1016\discretionary{/}{%
}{/}j\hspace{.1pt}\discretionary{.}{%
}{.}\hspace{.4pt}aei\hspace{.1pt}\discretionary{.}{%
}{.}\hspace{.4pt}2016\hspace{.1pt}\discretionary{.}{%
}{.}\hspace{.4pt}05\hspace{.1pt}\discretionary{.}{%
}{.}\hspace{.4pt}004}}


\bibitem{Xu2025}
T.~Xu, W.~Gu, K.~Ota, and S.~Hasegawa.
\newblock Development and evaluation of a low-jitter hand tracking system for improving typing efficiency in a virtual reality workspace.
\newblock {\em Multimodal Technologies and Interaction}, 9(1):4, 2025. doi: {{%
10\hspace{.1pt}\discretionary{.}{%
}{.}\hspace{.4pt}3390\discretionary{/}{%
}{/}mti9010004}}


\end{thebibliography}

\end{document}